\documentclass[letterpaper, 10 pt, conference]{ieeeconf}  %

\IEEEoverridecommandlockouts                              %

\usepackage{amsmath} %
\usepackage{amssymb}  %
\usepackage{booktabs}
\usepackage{multirow}
\usepackage{xcolor}
\usepackage{graphicx}
\usepackage{subcaption}
\usepackage{tikz}
\usepackage{algorithm}
\usepackage[noend]{algpseudocode}
\usepackage{makecell}

\title{\LARGE \bf
  Visibility-Aware Task and Motion Planning with Task-Centric Perception
}

\author{Sahit Chintalapudi, Xiaoyan Wang, Aryan Zoroufi, Aditya Agarwal, Nancy Kanwisher,\\
Tom\'{a}s Lozano-P\'{e}rez, and Leslie Pack Kaelbling%
\thanks{All authors are with the Massachusetts Institute of Technology, Cambridge, MA 02139, USA.
        Corresponding author: Sahit Chintalapudi, {\tt\small sahit@mit.edu}}%
}

\newcommand{\ourmethod}{\texttt{VA-TAMP}}
\newcommand{\para}[1]{\smallskip\noindent\textbf{#1.}\hspace{0.5em}}
\begin{document}

\maketitle
\thispagestyle{empty}
\pagestyle{empty}

\begin{abstract}
  Long-horizon manipulation under uncertainty remains a challenging problem in robotics. We present \ourmethod{}, an algorithm for Visibility-Aware Task and Motion Planning. \ourmethod{} extends existing deterministic task and motion planning solvers to partially observable object rearrangement tasks, while still being able to take advantage of the efficiency and compositional generalization over long horizons of the base solver. This extension interleaves sensing actions into a plan skeleton that assumed full observability, and uses a novel persistent object-based memory to fuse observations into object mesh hypotheses and an observed occupancy map. We evaluate \ourmethod{} on real-world tabletop rearrangement tasks which feature heavy occlusion and partial visibility and demonstrate that it outperforms baselines that do not perform active perception or have persistent memory.

\end{abstract}

\section{INTRODUCTION}
\label{sec:intro}
A longstanding challenge in robotic manipulation is completing long-horizon tasks in heavily occluded environments with unfamiliar objects. When object models and poses are known \textit{a priori}, Task and Motion Planning (TAMP) solvers~\cite{garrett2020pddlstream, shen2024differentiable} can efficiently compose manipulation skills to achieve a goal. Otherwise, object models can be estimated from an observation and passed to a TAMP solver~\cite{curtis2022long,shen2026tiptop}. These approaches assume that the initial observation models the scene accurately enough to find a goal-achieving plan.

\begin{figure}[t]
  \centering
  \begin{subfigure}[t]{0.49\linewidth}
    \includegraphics[width=\linewidth]{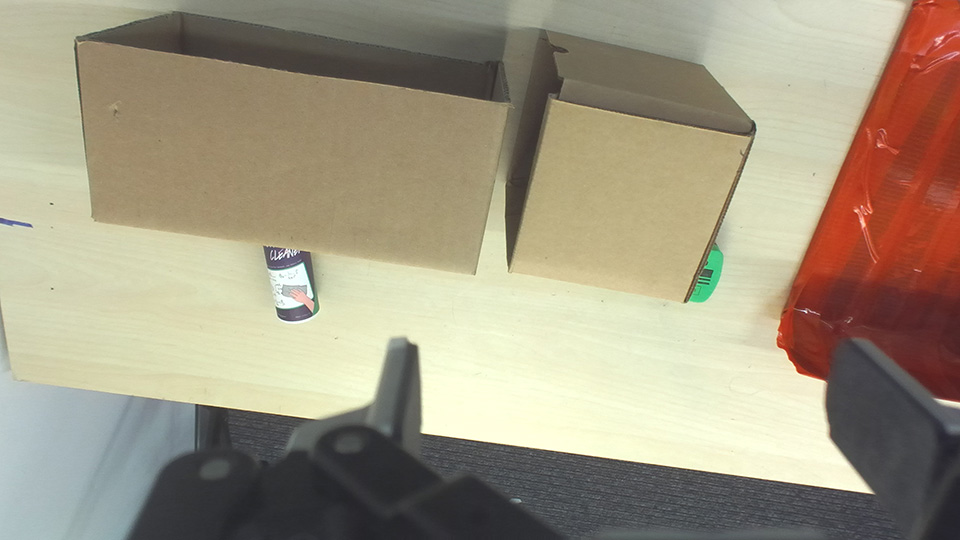}
    \caption{Initial wrist view. Goal: {\scriptsize\texttt{\{On(spray-bottle,\,red-tray), On(green-eraser,\,red-tray)\}}}}
    \label{fig:teaser-a}
  \end{subfigure}\hfill
  \begin{subfigure}[t]{0.49\linewidth}
    \includegraphics[width=\linewidth]{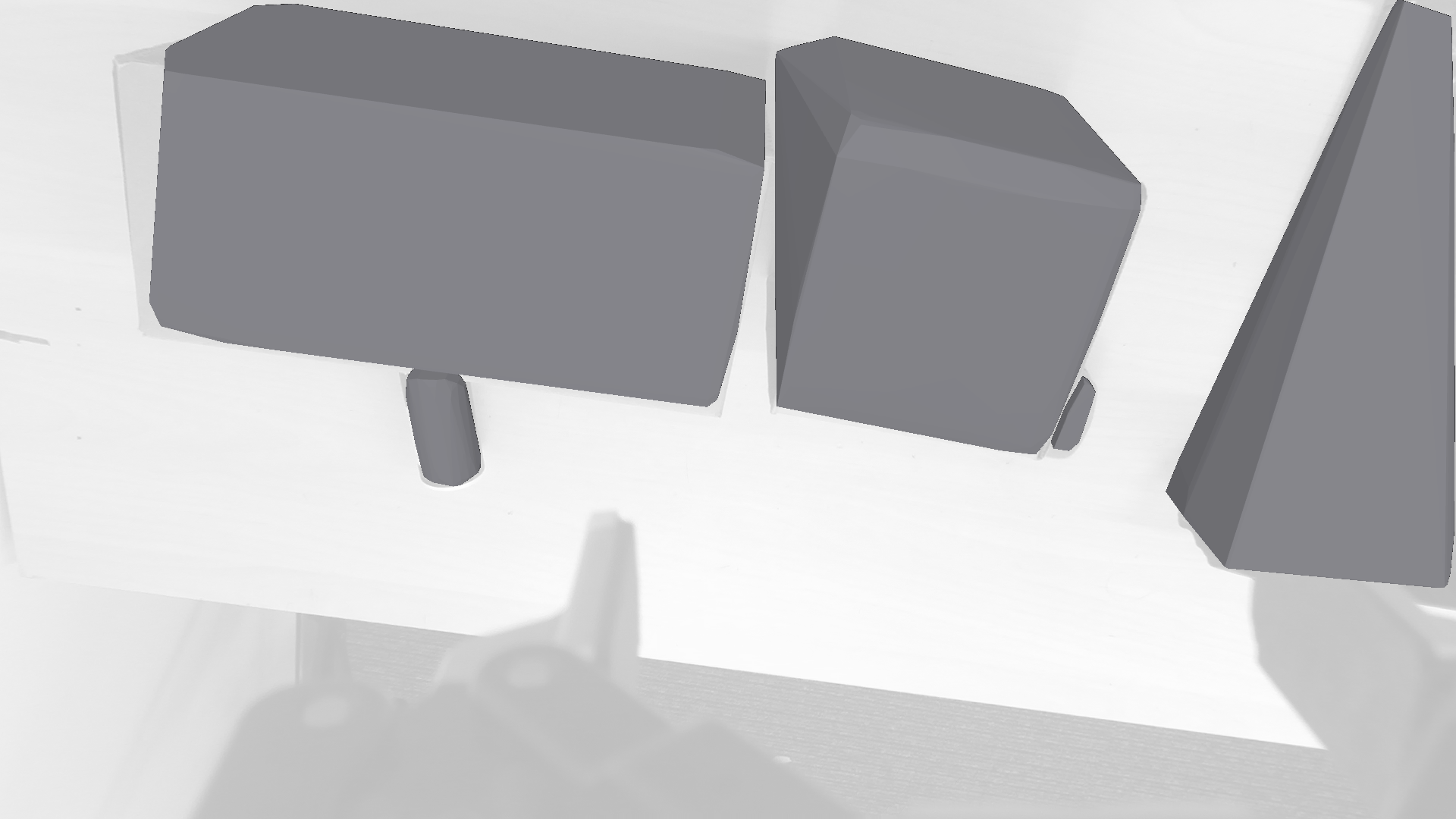}
    \caption{Initial scene estimate: a convex hull per detected object.}
    \label{fig:teaser-b}
  \end{subfigure}\\[4pt]
  \begin{subfigure}[t]{0.49\linewidth}
    \includegraphics[width=\linewidth]{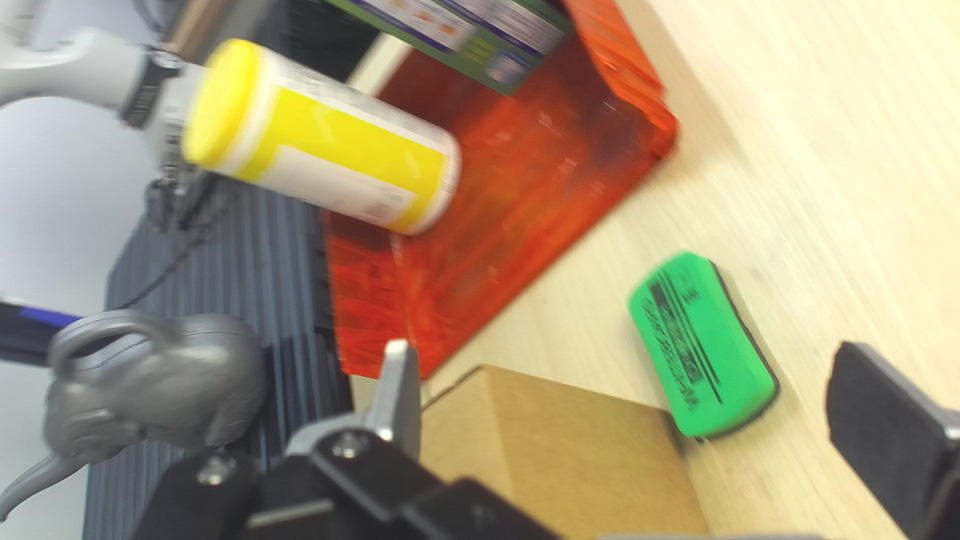}
    \caption{A planned wrist view reveals the full extent of the eraser.}
    \label{fig:teaser-c}
  \end{subfigure}\hfill
  \begin{subfigure}[t]{0.49\linewidth}
    \includegraphics[width=\linewidth]{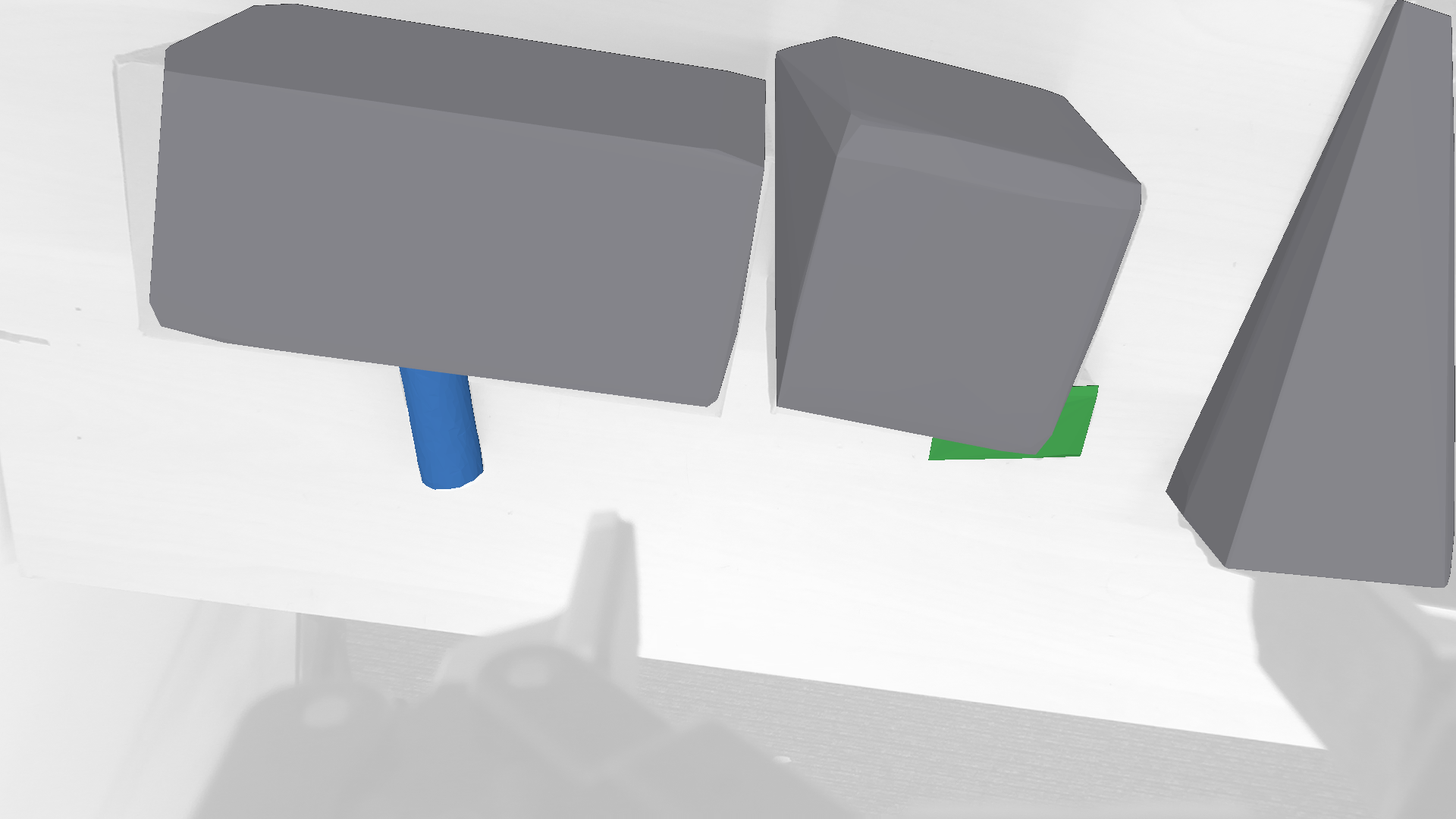}
    \caption{Task-relevant objects refined into completed shapes.}
    \label{fig:teaser-d}
  \end{subfigure}
  \caption{Visibility-Aware Task and Motion Planning}
  \label{fig:teaser}
\end{figure}

Consider a tabletop rearrangement task, such as the one illustrated in Figure~\ref{fig:teaser}, where the robot has to place the cleaning supplies on the red tray. In the initial observation, the full extent of the tray (which already has other cleaning supplies on it) is not visible to the robot, and the cleaning supplies that the robot has to manipulate are partially occluded by the cardboard boxes. This setting is partially observable; the robot does not know the full shape of any of the objects in the scene, and thus it must make a plan from raw observations which are noisy and incomplete.

\begin{figure*}[t]
  \centering
  \input{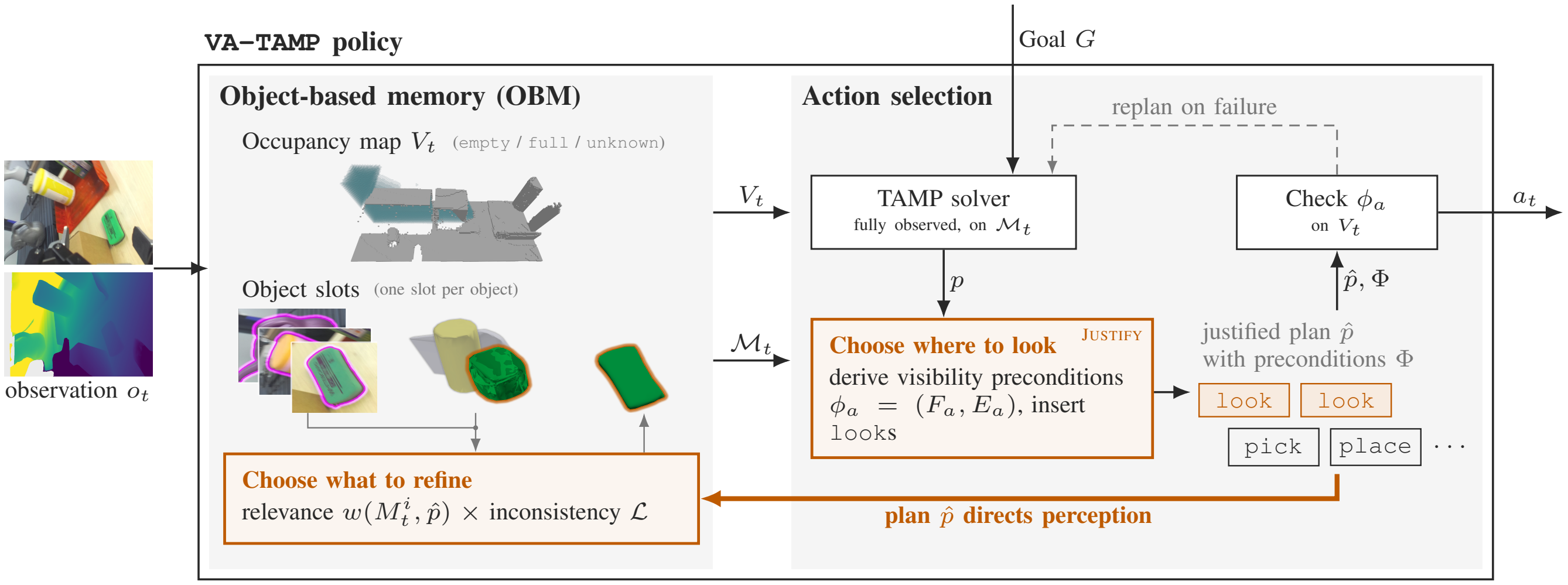}
  \caption{Overview of \ourmethod{}.
    RGB-D observations update an object-based memory (OBM), which maintains observed occupancy $V_t$ and object mesh hypotheses $\mathcal{M}_t$.
    Given a goal $G$, a fully observed TAMP solver computes a plan $p$ over $\mathcal{M}_t$.
    \textsc{Justify} uses $\mathcal{M}_t$ and $V_t$ to derive a set of visibility preconditions $\Phi$ and insert \texttt{look} actions to produce a justified plan $\hat p$.
    These preconditions are checked against $V_t$ to select the next action $a_t$ or trigger replanning.
  The plan also guides OBM refinement, prioritizing mesh hypotheses that are relevant to the plan and inconsistent with accumulated observations.}
  \label{fig:method-overview}
\end{figure*}

The naive approach to this problem, re-estimating and replanning with every new observation, suffers from two main shortcomings. First, any single observation is not sufficient to accurately model the scene. Second, TAMP solvers output strictly goal-oriented plans, meaning that they cannot direct the perception toward the task-relevant parts of the scene or remove occluding objects. In Figure~\ref{fig:teaser}, no single view shows the far edge of the tray, and a plan that only reaches for the eraser never looks behind the boxes that occlude it. On the other end of the spectrum, belief space planners explicitly model uncertainty in the environment state and search for plans by reasoning over full observation distributions obtained by each action~\cite{curtis2022task, bry2011rapidly, platt2016efficient, hauser2010randomized}. However, modeling geometric uncertainty requires reasoning over extremely high-dimensional distributions which quickly becomes intractable.

Our approach exploits the compositional generalization and efficiency of TAMP solvers while tracking object model accuracy and unobserved regions of space. \ourmethod{} augments off-the-shelf TAMP solvers with action constraints requiring the robot to verify its object models before interacting with the objects. These constraints induce information gathering actions that provide observations for refining the models.

In order to build and refine our scene representation as we observe more of the scene and objects move, we define an \textit{Object-Based Memory} module (OBM). The OBM collects a history of observations and maintains two complementary representations of the scene. The first representation is an occupancy grid, which accumulates depth data and tracks the regions of space that have not yet been observed. The second representation is a set of objects that are believed to exist in the scene after being detected in individual observations. The OBM associates these detections on a per-object basis and leverages a generative image-to-3D reconstruction model~\cite{chen2026sam,zadaianchuk2026recgen} to hypothesize a complete 3D mesh geometry from partial RGBD observations. The OBM is resource-rational and task-centric: it focuses its computation time on objects that the planner has decided are relevant to the task. In the example, the eraser and tray are refined into complete meshes (Figure~\ref{fig:teaser-d}) while the cardboard boxes are left as the coarse hulls from the first frame (Figure~\ref{fig:teaser-b}).

Given a set of object meshes, we can leverage existing classical TAMP solvers to generate a plan skeleton: an optimistic plan that assumes the current object meshes are accurate and is entirely goal-oriented. To guard against this optimism, we require that every plan be \textit{justified}. This means that any time the plan requires the robot to move through space, it must have observed that the space is collision free. Any time the plan requires the robot to grasp an object at a given pose, it must have observed that the object surface exists in between the fingers at that pose. To justify a plan, the planner has to find motions that move the camera to viewpoints where the desired regions are visible. The initial hull of the green eraser (Figure~\ref{fig:teaser-b}) does not justify a grasp; the planner inserts the wrist view in Figure~\ref{fig:teaser-c}, which reveals the eraser's full extent and lets the grasp proceed.

\ourmethod{} extends fully observed TAMP to partially observed domains by coupling object-based memory with scene-model verification. The memory supplies object meshes and an occupancy grid, and the planner constructs a goal-reaching plan that also validates the meshes. As observations arrive, the remaining plan guides refinement of task-relevant objects. Figure~\ref{fig:method-overview} illustrates this coupling. Our contributions are as follows.

\para{\ourmethod: An algorithm for Visibility-Aware Task and Motion Planning} \ourmethod{} augments classical TAMP solvers with a persistent memory system and task-driven information gathering behavior. We show that this method outperforms systems which don't perform active perception or do not associate object detections across observations.

\para{A task-centric, resource-rational OBM} Accurately modeling a scene is a computationally expensive task, with state-of-the-art completion models taking 4-5 seconds per object in a single frame. Rather than completing every object in every frame, the OBM prioritizes reconstruction on objects that are task-relevant and do not yet have a good model.

\para{Efficient information gathering in partially observable problems by leveraging a fully observed TAMP solver} We use the result of an existing TAMP solver to define which regions of space the robot interacts with, either as motion in assumed free space or contact with assumed object surfaces. We make the observation of these regions of space a precondition to acting, forcing the robot to collect more observations and update its models as needed.

\section{RELATED WORK}
\subsection{Visibility-Aware Planning}

Goretkin et al.~\cite{goretkin2018look} formulate visibility-aware motion planning by requiring a robot to observe a region as empty before moving through it. Muguira-Iturralde et al.~\cite{muguira2023visibility} extend this formulation to allow the robot to move occluders on the trajectory. Ashur and Sintov~\cite{ashur2026sampling} instead treat visibility as an objective and seek configurations with a clear line of sight to a target. Next-best-view planners~\cite{Lauri_RAL2020, delmerico2018comparison} select views to collect information efficiently, but are generally task-agnostic. Our method uses visibility constraints to verify both free space along planned motions and object surfaces at planned grasp contacts. The resulting observations also guide refinement of the object geometry.

\subsection{Task and Motion Planning}

Task and Motion Planning (TAMP) is a well-studied problem formulation, and here we focus on TAMP approaches that operate without access to the ground-truth state.
Curtis et al.~\cite{curtis2022long} and Shen et al.~\cite{shen2026tiptop} estimate the scene from an initial observation and treat this estimate as ground truth for a deterministic TAMP solver. \ourmethod{} builds on the same class of solvers, but verifies and updates the estimate as the plan executes.

Another class of TAMP approaches plan in belief space using determinized observation models~\cite{Menell2015Modular, garrett2020online} or by weighting observation outcomes by their likelihood~\cite{kaelbling2013integrated}. These approaches replan when new observations invalidate the current plan. Curtis et al.~\cite{curtis2024partially} avoid determinization by learning a belief-space MDP from simulated operator rollouts and solving that MDP. Our OBM instead keeps a single mesh hypothesis per object at a time and evaluates how well it explains the accumulated observations, without explicitly maintaining a distribution over possible geometries.

Recent work has also used Vision-Language Models (VLMs) to handle uncertainty in TAMP problems. Zhao et al.~\cite{zhao2025seeing} use a VLM to classify predicates as known or unknown and plan observation actions to resolve the unknown ones. Tanaka et al.~\cite{tanaka2026evidence} and Oloo et al.~\cite{oloo2026robot} use VLMs to propose observation actions and decide whether the robot has enough information to act. These methods use VLMs to reason about symbolic uncertainty and select observations. \ourmethod{} instead addresses continuous geometric uncertainty by deriving observation targets from planned motions and contacts.

\subsection{Scene Representation}

Mitash et al.~\cite{mitash2020task} choose each object's representation according to its role in the plan. Objects to be grasped are represented by minimal alpha shapes, which restrict grasps to observed surfaces. Objects to be avoided are represented by conservative hulls that include their unobserved shadows. Our OBM instead updates object meshes throughout the episode, using the observations to evaluate each mesh and the plan to prioritize refinement.

To fuse observations over time, Gu et al.~\cite{gu2024conceptgraphs} build a persistent 3D scene graph by associating objects using geometric overlap and visual feature similarity. Du et al.~\cite{du2022learning} learn an object-based memory that updates each slot's shape estimate as new observations arrive, including in scenes where objects move.

Task-agnostic reconstruction methods use generative image-to-3D models to estimate complete object meshes and poses from sparse views~\cite{agarwal2025scenecomplete, huang2026simulation, yu2026picasso}. Zadaianchuk et al.~\cite{zadaianchuk2026recgen} jointly estimate object pose, scale, and geometry. Our OBM instead updates a persistent scene representation as observations arrive and objects move, while allocating reconstruction effort according to the current manipulation plan.

\section{Partially Observable Object Rearrangement}
\subsection{Problem Statement}
We formulate long-horizon manipulation under partial observability as a Partially Observable Object Rearrangement (POOR) problem, an instance of a stochastic shortest-path problem on a belief space MDP whose underlying state is defined in terms of posed meshes.

The robot manipulator has a set of fully observed joint angles $q$ and is embedded in an environment with an unknown number $N^o$ of objects. Each object in the object set $\mathcal{M}$ is a posed mesh $M^i = (m^i, X^i) \in \mathcal{M}$, where $m^i$ is a tuple of surface vertices and faces, and $X^i$ is the pose of the mesh centroid relative to a known world frame. Neither the object geometries nor their poses are provided a priori, and both must be inferred from observations. The state space $\mathcal{S}$ of a POOR problem is the space of possible sets of geometries and robot configurations.

At every step, the robot takes a temporally extended parameterized action $a \in \mathcal{A}$ (e.g., \texttt{Pick$(X^{\text{hand}})$}, \texttt{Place$(X^{\text{hand}})$}, \texttt{Look$(X^{\text{hand}})$}), which returns an observation sequence $o_{1:T} \in \mathcal{O}$. Each observation $o_t = (q_t, \mathcal{I}_t)$ contains the robot's joint angles $q_t$ and an RGBD image $\mathcal{I}_t$ captured at time $t$. We assume access to a transition function $\mathcal{T}: \mathcal{S} \times \mathcal{A} \rightarrow \mathcal{S}$ which describes how actions change the state.

The robot is given an initial observation $o_0$ and a goal $G$, which is a function on the history of observations that is true when a desired property holds in the world. For example, given a text label of an object $l_0$ and desired support surface $l_s$, the robot may try to achieve the goal $\exists x, y \; s.t. \;  \texttt{Is}(x, l_0) \land \texttt{Is}(y, l_s) \land \texttt{On}(x, y)$. The solution to a POOR problem is a policy $\pi: \mathcal{O} \rightarrow \mathcal{A}$ which maps observations to actions, is executed starting from the initial observation, and is evaluated by computing the cost of the actions taken to move from the initial state to a goal-satisfying state.

\subsection{Overall Approach}
Computing a full policy mapping from any observation to the correct action completely offline is impractical due to the complexity of the POOR problem. Instead, \ourmethod{} takes a model-predictive control style approach to constructing a policy. We combine a state estimator which computes models of the scene conditioned on observations with an action-selection module which selects an action conditioned on the goal and current scene models. On execution of the action, the policy receives a new observation, which it uses to update its model and continue execution or replan.

This approach to constructing the policy is illustrated in Algorithm~\ref{alg:vatamp-full}. We initialize the \emph{Object-Based Memory} from the initial observation (line 1) and use this to estimate a hypothesized set of meshes in the world and a voxel occupancy grid (line 3). The action selection (lines 4-5) starts by solving a fully observed TAMP problem assuming $\mathcal{M}$ is the true state to get a plan backbone $p$. The \texttt{JUSTIFY} procedure derives observation-conditioned preconditions for manipulation actions and then interleaves sensing actions into the plan to verify the preconditions to get the justified plan $\hat{p}$. If the preconditions hold in the true scene, we execute the actions in the plan and receive new observations, which are used to update the OBM. If the preconditions do not hold, the OBM is queried for a fresh scene model to plan against.

\begin{algorithm}[t]
  \caption{\ourmethod}
  \label{alg:vatamp-full}
  \begin{algorithmic}[1]
    \Require goal $G$, initial observation $o_0$, robot pose $q_0$
    \State $\mathcal{B} \gets \Call{Init-OBM}{o_0}$
    \While{$\neg G$}
    \State $\mathcal{M}, V\ \gets \Call{Estimate-Scene}{\mathcal{B}}$
    \State $p \gets \Call{Solve-TAMP}{q_0, \mathcal{M}, G}$
    \State $\hat{p}, \Phi \gets \Call{Justify}{p, \mathcal{M}, V}$
    \For{$a, \phi_a \in \hat{p}, \Phi$}
    \If{$\phi_a(V_t, \mathcal{M}_t)$}
    \State $o_{1:t}, q_t \gets \Call{Execute}{a}$
    \State $\Call{Update}{\mathcal{B}, o_{1:t}, \hat{p}}$
    \Else
    \State \textbf{break} \Comment{replan}
    \EndIf
    \EndFor
    \EndWhile
  \end{algorithmic}
\end{algorithm}

In Section~\ref{sec:obm_method}, we describe the \emph{resource-rational} OBM, which trades perception computation for utility in downstream planning. In Section~\ref{sec:vatamp_method} we define the \texttt{JUSTIFY} procedure used to derive action preconditions and sensing actions.

\section{Task-Centric Object-Based Memory}
\label{sec:obm_method}
\subsection{Problem Statement}
The OBM is responsible for computing up-to-date representations of the scene to be used by a downstream planner. Formally, the OBM takes as input an observation history $o_{1:t}$ and the robot's current plan $\hat{p}$. It returns a collection of posed surface meshes $\mathcal{M}_t$ and an occupancy map $V_t$. Generally, occupancy maps associate every voxel with a probability of that voxel being occupied~\cite{moravec1985high, elfes1989using}. Our occupancy map is a discretized special case of this and takes the form
\begin{align*}
  V^r(x, y, z | o_{1:t}) \in \{\texttt{empty}, \texttt{full}, \texttt{unknown}\}.
\end{align*}

Here $(x, y, z)$ is any point in the bounded workspace, expressed in the world frame, and $V^r$ returns the label of the voxel of side length $r$ containing it.

Estimating $\mathcal{M}_t$ requires hypotheses about unobserved object surfaces. Thus, we assume access to a generative shape model $\mathcal{G}$ that maps RGBD observations of an object to a posed mesh hypothesis for the full object. Querying $\mathcal{G}$ is slow relative to the observation frequency, so the OBM makes at most one query per \textit{perception step}, running in parallel with robot execution. Estimating $\mathcal{M}_t$ is therefore a sequential decision problem: at each step, the OBM chooses which object to refine and whether the returned mesh should replace its current hypothesis.

The OBM is evaluated on the quality of its mesh representation $\mathcal{M}_t$. Let $M_t^i$ denote the posed mesh estimate of object $i$ at perception step $t$, and let $M^i$ denote its true posed mesh. Given the plan, the OBM's goal is to minimize
\begin{align}
  J(\mathcal{M}_{0:T}) &= \sum_{t=0}^{T} \sum_{i=1}^{N^o} w(M_t^i, \hat{p}_t)\, e(M_t^i, M^i), \nonumber \\
  \label{eq:taskweight}
  w(M_t^i, \hat{p}_t) &= \exp\!\Big(-\beta \min_{x \in M_t^i,\; y \in \tau_{ee}(\hat{p}_t)} \|x - y\|\Big).
\end{align}

Here $e$ measures the discrepancy between the estimated mesh and the true object, and $\tau_{ee}(\hat{p}_t)$ is the end-effector path swept by the current plan. Since the true meshes are unavailable at runtime, the OBM minimizes a surrogate, the cost of each mesh under its observations, defined in the following section.

\subsection{Memory Representation and Updates}\label{sec:obm_approach}
Our OBM maintains a set of \textit{object slots}, each representing a hypothesized object through its observation history and current 3D geometry estimate. An object detection is $d_t^i = (o_t, \ell_t)$, where $o_t$ is an observation containing the object and $\ell_t$ is a binary mask of its image pixels.
We define a slot as
\begin{align*}
  s_t^i = (M_t^i, \mathcal{D}_t^i, \mathcal{P}_t^i).
\end{align*}
where $\mathcal{D}_t^i$ is the set of detections attributed to object $i$, and $\mathcal{P}_t^i$ contains the accumulated world-frame points obtained by back-projecting the depth pixels assigned to object $i$ in these detections.

The object slots are updated based on each observation, and these slots are then used to update the mesh-based representation and occupancy-based representation. From every observation, a set of detections is computed. These detections are then either associated with an existing object slot, or initialize a new object slot. The OBM updates its mesh-based representation by choosing a slot to run the completion model on and updates its occupancy-based representation by aggregating the depth images and updating regions of space based on object rearrangement.

\para{Acquiring and Associating Detections} For each observation $o_t$, we query a VLM for bounding boxes of objects on the table, then use a segmentation model to refine each box into an object mask. The resulting detections $d_t^i$ are matched to existing slots or treated as novel objects.

To assign detections to the appropriate object slot, we follow a similar strategy to ConceptGraphs~\cite{gu2024conceptgraphs}. A new detection $d = (o, \ell)$ yields a set of world-frame points $\mathcal{P}(d)$, obtained by back-projecting the depth pixels under the mask $\ell$. It also yields a visual feature $f(d)$, the mean of the dense DINO features~\cite{oquab2023dinov2} over the same pixels. The corresponding quantities for a slot $s_t^i$ are its accumulated points $\mathcal{P}_t^i$ and the mean feature over its detections, $f(s_t^i) = \frac{1}{|\mathcal{D}_t^i|}\sum_{d' \in \mathcal{D}_t^i} f(d')$.
We score a detection against a slot as
\begin{align*}
  \mathrm{score}(d, s_t^i) = \tfrac{1}{2}\big(\, g(d, s_t^i) + \cos(f(d), f(s_t^i)) \,\big),
  \\
  g(d, s_t^i) = \max\!\big(\, c(\mathcal{P}(d) \!\to\! \mathcal{P}_t^i),\; c(\mathcal{P}_t^i \!\to\! \mathcal{P}(d)) \,\big),
\end{align*}
where $c(P \!\to\! Q)$ is the fraction of points in $P$ within $\delta$ of some
point in $Q$. A slot is a candidate only if $g \ge \tau_g$ and
$\cos(f(d), f(s_t^i)) \ge \tau_f$ for a specified geometric similarity threshold $\tau_g$ and visual similarity threshold $\tau_f$. Within each observation, detections are processed in descending order of their best candidate score. Each detection joins the highest-scoring candidate slot that has not already been assigned, appending
$d$ to $\mathcal{D}_t^i$ and $\mathcal{P}(d)$ to $\mathcal{P}_t^i$, or seeds
a new slot if there are none.

A detection $d$ seeds a new slot by instantiating $s_t^i = (M_t^i, \{d\}, \mathcal{P}(d))$, with $M_t^i$ denoting the convex hull of $\mathcal{P}(d)$.

\para{Updating the Mesh Estimate}
At every perception step, the OBM chooses which object model to update. Because the true geometries in the scene are unknown, we define a surrogate cost which measures how inconsistent a candidate posed mesh $M'$ is with the object detections $\mathcal{D}_t^i$ and points $\mathcal{P}_t^i$. The cost $\mathcal{L}$ is a negative log-likelihood given by
\begin{align*}
  \mathcal{L}(M', \mathcal{P}, \mathcal{D}) = {}& \lambda_{DDM}\,\mathcal{L}_{DDM}(M', \mathcal{P})  \\
  & + \lambda_{empty}\,\mathcal{L}_{empty}(M', \mathcal{D})  \\
  & + \lambda_{recall}\,\mathcal{L}_{recall}(M', \mathcal{D}),
\end{align*}
where $\mathcal{L}_{DDM}$ is the directional distance metric proposed by~\cite{ren2024measuring} and is a correspondence-free approach to measuring the alignment between observed points and the mesh. $\mathcal{L}_{empty}$ and $\mathcal{L}_{recall}$ are negative log-likelihoods comparing the mesh rendered at the proposed scale, rotation, and translation to the observations, derived from the Bayes3D perception system~\cite{gothoskar2023bayes3d}.

The OBM then greedily picks the slot whose current estimate $M_t^i$ is the most inconsistent with its evidence, weighted by task relevance,
\begin{equation*}
  i^{*} = \arg\max_{i}\; w(M_t^i, \hat{p}_t)\, \mathcal{L}(M_t^i, \mathcal{P}_t^i, \mathcal{D}_t^i),
\end{equation*}
where $w$ is the task-relevance weight defined in Equation~\ref{eq:taskweight}.

In practice, the majority of current image-to-3D reconstruction models can only be conditioned on a fixed number of views, and not an arbitrarily long list of detections. Thus, after selecting $i^*$, the OBM selects the detection $d \in \mathcal{D}_t^{i^*}$ that reveals the largest fraction of object $i^*$ and uses this as the input to $\mathcal{G}$. We then evaluate $\mathcal{L}_\text{cand} = \mathcal{L}(\mathcal{G}(d), \mathcal{P}_t^{i^*}, \mathcal{D}_t^{i^*})$ as the cost of the model output and $\mathcal{L}_\text{curr} = \mathcal{L}(M_t^{i^*}, \mathcal{P}_t^{i^*}, \mathcal{D}_t^{i^*})$ as the cost of the current slot estimate. If $\mathcal{L}_{\text{cand}} < \mathcal{L}_{\text{curr}}$, then the slot mesh $M_t^{i^*}$ is replaced with $\mathcal{G}(d)$.

\para{Updating the Occupancy Grid} The voxel grid $V$ is instantiated from the first observation $o_0$ and updated in real time as new observations arrive. We represent the voxel grid in memory as an octree~\cite{hornung13auro} which stores an estimate of the log-odds of each cell being occupied and supports efficient Bayesian updates. A cell is marked \texttt{unknown} if it is not observed in any depth image, \texttt{empty} if the probability of the cell being occupied is less than 0.5, and \texttt{full} otherwise.

Picking up and placing an object requires an additional update to $V$ because grasped objects occlude much of the visible area of a wrist mounted camera. When the robot picks up object $i$, the cells containing the corresponding slot's points $\mathcal{P}_t^i$ at the object's initial pose are marked as \texttt{unknown} in $V$. The cells are marked \texttt{unknown} and not \texttt{empty} because noise in the depth measurements and segmentation models means that $\mathcal{P}_t^i$ may contain points on the surfaces of objects that neighbor or support object $i$, so marking the cells \texttt{unknown} more accurately captures the uncertainty about the cell state after grasping.

\section{Plan Justification}
\label{sec:vatamp_method}
\subsection{Problem Statement}
The input to plan justification is a plan $p = (a_1, \ldots, a_n)$ from a TAMP solver that treated the OBM's mesh hypotheses $\mathcal{M}_t$ as ground truth. Each action $a_u$ has a target robot pose $X_{a_u}$, a joint-space trajectory $\tau$ reaching that pose, a target object $o^u$, and a constraint set \texttt{con($a_u, \mathcal{M}_t$)}. These constraints must all hold for the action to be valid; for example, $X_{a_u}$ must not collide with any object in $\mathcal{M}_t$.

The output is a \textit{justified} plan $\hat{p}$, which interleaves \texttt{Look} actions with the original manipulation actions in $p$, and a set of \textit{visibility preconditions} $\Phi = \{\phi_a\}_a$. These preconditions are derived from $p$ and take the form
\begin{align}
  \label{eq:pcf}
  \phi_a(V_t, \texttt{con}(a, \mathcal{M}_t)) \in \{\text{true}, \text{false}\}
\end{align}
and are tested before the action $a$ executes. The precondition $\phi_a$ holds when the constraints on the hypothesized meshes $\mathcal{M}_t$ are supported by the observed occupancy grid.

A justified plan should be \textit{valid}: before each manipulation action $a_u$, the preceding \texttt{Look} actions should have labeled every voxel its precondition depends on. It should also be \textit{efficient}, avoiding observations that increase the original plan's cost without validating or helping refine a task-relevant object.

\subsection{Constructing Justified Plans}
\para{The Justified Plan}
Pseudocode for \textsc{Justify} is given in Algorithm~\ref{alg:justify}. The plan $p$ returned by the TAMP solver in Algorithm~\ref{alg:vatamp-full} is \textit{unjustified}: it is valid against $\mathcal{M}_t$, but $V_t$ may not contain observations that support every action in $p$. \textsc{Justify} assumes that the plan consists of a sequence of \texttt{Pick}, \texttt{Place} pairs. It starts by deriving the visibility preconditions for each of these actions (lines 3--4), then inserts \texttt{look} actions into the original plan (lines 5--6), and returns the \textit{justified} plan $\hat{p}$ together with the set of preconditions $\Phi$ that Algorithm~\ref{alg:vatamp-full} checks before executing each manipulation action.

\begin{algorithm}[t]
  \caption{JUSTIFY}
  \label{alg:justify}
  \begin{algorithmic}[1]
    \Require unjustified plan $p$, mesh set $\mathcal{M}_t$, occupancy map $V_t$
    \State $\hat{p} \gets ()$;\quad $\Phi \gets \emptyset$
    \For{$(\texttt{Pick}(o^u, X),\ \texttt{Place}(o^u, X')) \in p$}
    \State $\phi_{\mathrm{pick}} \gets \Call{DerivePre}{\texttt{Pick}(o^u, X), \mathcal{M}_t}$
    \State $\phi_{\mathrm{place}} \gets \Call{DerivePre}{\texttt{Place}(o^u, X'), \mathcal{M}_t}$
    \State $\hat{p} \gets \hat{p} \,\Vert\, \big(\texttt{Look}(\phi_{\mathrm{pick}}),\ \texttt{Look}(\phi_{\mathrm{place}})\big)$
    \State $\hat{p} \gets \hat{p} \,\Vert\, \big(\texttt{Pick}(X),\ \texttt{Place}(X')\big)$
    \State $\Phi \gets \Phi \cup \{\phi_{\mathrm{pick}}, \phi_{\mathrm{place}}\}$
    \EndFor
    \State \Return $\hat{p}, \Phi$
  \end{algorithmic}
\end{algorithm}

Looks  are inserted by a fixed rule. For each \texttt{pick} and \texttt{place} pair in $p$, a \texttt{look} for the pick's precondition and another for the place's precondition are both inserted immediately before the \texttt{pick}. The place's look is not deferred until after the pick because, once the robot grasps something, the hand is no longer free and the wrist camera cannot collect new observations. Algorithm~\ref{alg:vatamp-full} then executes $\hat{p}$ in order, checking each precondition in $\Phi$ against the current $V_t$ before its action runs and returning to the OBM for a fresh scene estimate when a check fails.

\begin{figure}[t]
  \input{figures/pcf/fig_pcf_body}
  \caption{A precondition for \texttt{pick}$(\text{eraser}, g, q)$. (a) The precondition computed from an estimated $\mathcal{M}_t$. The finger contacts $F_a$ (pink) must read \texttt{full}, and the swept volume of the gripper along the grasp trajectory $E_a$ must read \texttt{empty}. The blue cells are known to be empty, the orange cells are unknown. (b) After the look, the required cells are observed and $\phi_a$ holds.}
  \label{fig:pcf}
  \vspace{-14pt}
\end{figure}

\para{Deriving Visibility Preconditions}
\label{sec:pcf}
Every visibility precondition $\phi_a$ of Equation~\ref{eq:pcf} is a tuple of point sets
\begin{align*}
  \phi_a = \big(F_a, E_a\big), \qquad F_a, E_a \subset \mathbb{R}^3,
\end{align*}
where every point in $F_a$ must be labeled \texttt{full} and every point in $E_a$ must be labeled \texttt{empty}. Figure~\ref{fig:pcf} shows the precondition of a \texttt{pick} action derived while manipulating the green eraser in the motivating example of Section~\ref{sec:intro}, before and after the \texttt{look} that satisfies it.

\textsc{DerivePre} computes $\phi_a$ from the constraint set on the meshes. The solver represents a \texttt{Pick} as \texttt{pick}$(o^u, X^{{\text{grasp}}}, \tau)$, with an object $u$ and a grasp pose $X$ relative to $o^u$, and a \texttt{Place} as \texttt{place}$(o^u, X^{\mathrm{place}}, \tau)$, with a placement pose $X^{\mathrm{place}}$ in the world frame.
\begin{itemize}
  \item \texttt{pick}$(o^u, X^{{\text{grasp}}}, \tau)$: the antipodal constraint on $X^{{\text{grasp}}}$~\cite{mahler2017dex}  depends on the surface of $o^u$ at the two finger contacts, so $F_a$ contains the points on $o^u$ where the finger pads close. The collision-free constraint is on the volume swept by the arm along $\tau$ while approaching the grasp, so $E_a$ contains samples of this volume.
  \item \texttt{place}$(o^u, X^{\mathrm{place}}, \tau)$: the support constraint on $X^{\mathrm{place}}$ starts by defining the ``footprint'' of $o^u$ as the projection of the volume $o^u$ onto a plane at the lowest point of $o^u$. $F_a$ contains a column of points below each cell of the footprint, satisfied when any point in the column is labeled \texttt{full}, which would imply the footprint is supported. The collision-free constraint depends on the footprint at $X^{\mathrm{place}}$, the volume swept by the held object as the robot follows $\tau$, and the volume swept by the arm, so $E_a$ contains samples of all three.
\end{itemize}

The visibility precondition $\phi_a$ holds when a fixed fraction of the constrained points have been observed and have the required label. Preconditions are derived once from the plan with respect to $\mathcal{M}_t$ and re-derived whenever $\mathcal{M}_t$ changes, since a new mesh hypothesis updates the contacts, footprint and swept volumes that define them.

\para{The \texttt{look} Operator}
Satisfying a visibility precondition requires new observations. The \texttt{Look} action is parameterized as \texttt{look}$(\phi_a, X)$, taking a precondition and a sequence of camera poses $X = (X_1, \ldots, X_m)$. A single pose is insufficient when the points of a precondition do not fit in one field of view, or are occluded from one direction and visible from another. Its precondition is that the hand is empty so that the wrist-mounted camera is unobstructed, its effect is that the points of $\phi_a$ have been labeled in $V_t$, and its constraint is that each pose in $X$ is reachable and collision-free against $\mathcal{M}_t$.

The sampler for $X$ is a greedy set cover over the subset of $F_a \cup E_a$ that is currently labeled \texttt{unknown} in $V_t$. We assume a camera model that predicts, for a candidate pose, which points fall in the field of view and are unoccluded by the meshes in $\mathcal{M}_t$. Each round selects an \texttt{unknown} point and samples candidate camera poses on the surface of a sphere around that point, oriented to look at the point. Samples are discarded if they are kinematically infeasible, are predicted to be occluded by other objects in the scene, or move the robot through points not labeled \texttt{empty} in $V_t$. Among the surviving samples, we keep the one whose field of view contains the most points in the precondition that are still \texttt{unknown}. Sampling continues until the chosen poses are expected to cover a fixed fraction of the precondition's required points or the view budget is exhausted. If no cover can be found, the precondition is marked unobservable; its precondition then fails at execution time and Algorithm~\ref{alg:vatamp-full} replans. To execute the look skill, a collision-free trajectory is planned that sequentially visits each viewpoint.

Because the underlying TAMP solver assumes full observability, it will not output actions that improve the visibility of the scene, although modifying the TAMP solver to move visual occluders is feasible. Therefore the \texttt{Look} actions assume that there is some viewpoint from which the region $\phi_a$ can be viewed, rather than planning actions to increase the visibility. In order to construct a solvable problem for the fully observed TAMP planner, we assume that the initial observation contains every object referenced in the goal statement. We also assume that while \texttt{Look} actions return noisy observations, they are not so noisy that collecting more observations worsens the quality of $\mathcal{M}_t$, and $V_t$. That is, collecting more observations should generally yield a better model of the world, not a worse one.

\section{EXPERIMENTS}
Partially observable object rearrangement is an ubiquitious but surprisingly under-studied problem setting and, as such, there are no directly comparable existing methods. In this section, we explore the effectiveness of \ourmethod{} in three scenarios of increasing complexity that were specifically designed to stress different aspects of the problem. We are interested in understanding when the complexity of a system like \ourmethod{} is justified relative to simpler methods, such as TiPToP, as well as to ablations of \ourmethod{}. Our ablations and tasks are designed to explore the following questions:

\textbf{Q1:} Do planned views derived from the current scene estimate generate useful targets for active perception?

\textbf{Q2:} Does a task-centric approach to model refinement improve downstream planning performance?

We answer \textbf{Q1} by defining the visibility-unaware ablation \texttt{VUTAMP}, which does not actively plan observation actions or derive visibility preconditions for replanning. Instead, the memory provides a set of mesh estimates used for fully observed planning, and replans are only triggered if a memory update invalidates the current plan by introducing a collision (this happens when the memory update grows an obstacle, and a nearby collision-free configuration now collides with the grown obstacle).

We answer \textbf{Q2} by defining the \texttt{VUTAMP-UNIFORM} ablation. This ablation follows the same planning and replanning loop as \texttt{VUTAMP}, but the update strategy for the OBM is task-agnostic. Every object in the scene is considered equally relevant and the OBM focuses on refining whichever slot has the highest error with respect to the observation sequence.

Finally, we compare against TiPToP~\cite{shen2026tiptop}, an open-loop model-based planning system that composes foundation models to estimate the initial scene. TiPToP forms initial conservative estimates of the scene and executes the generated plan open-loop, without updating on new observations.

\subsection{Experimental Setup}
\begin{figure}[t]
  \centering
  \begin{subfigure}[t]{0.32\linewidth}
    \includegraphics[width=\linewidth]{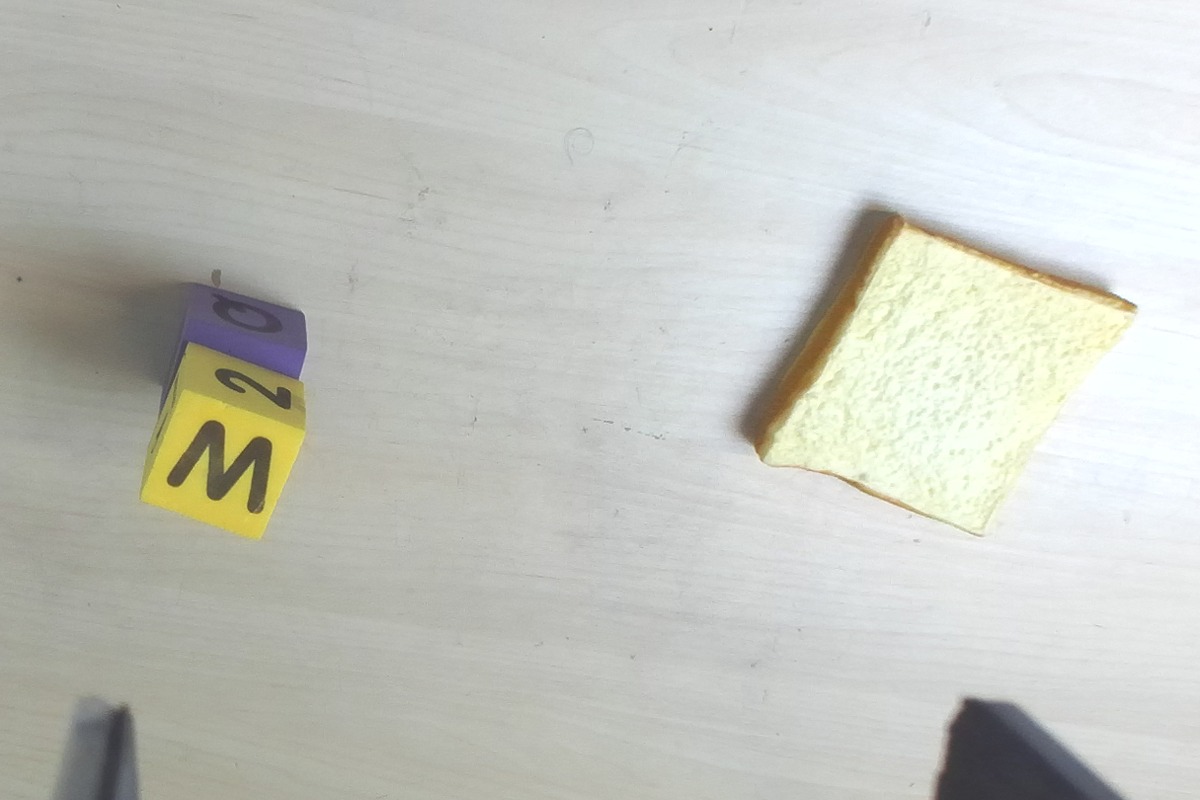}
    \caption{Cubes on Bread}
    \label{fig:scenes-cubes}
  \end{subfigure}\hfill
  \begin{subfigure}[t]{0.32\linewidth}
    \includegraphics[width=\linewidth]{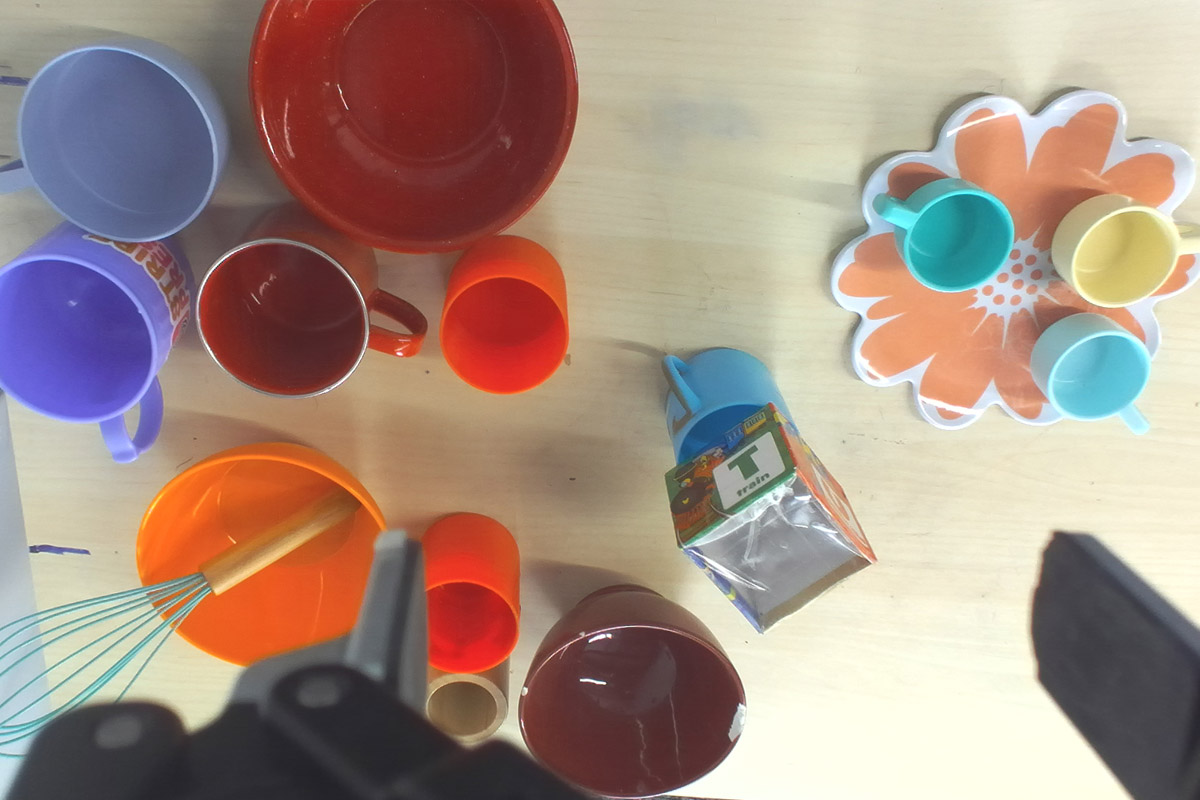}
    \caption{Teacup on Plate}
    \label{fig:scenes-teacups}
  \end{subfigure}\hfill
  \begin{subfigure}[t]{0.32\linewidth}
    \includegraphics[width=\linewidth]{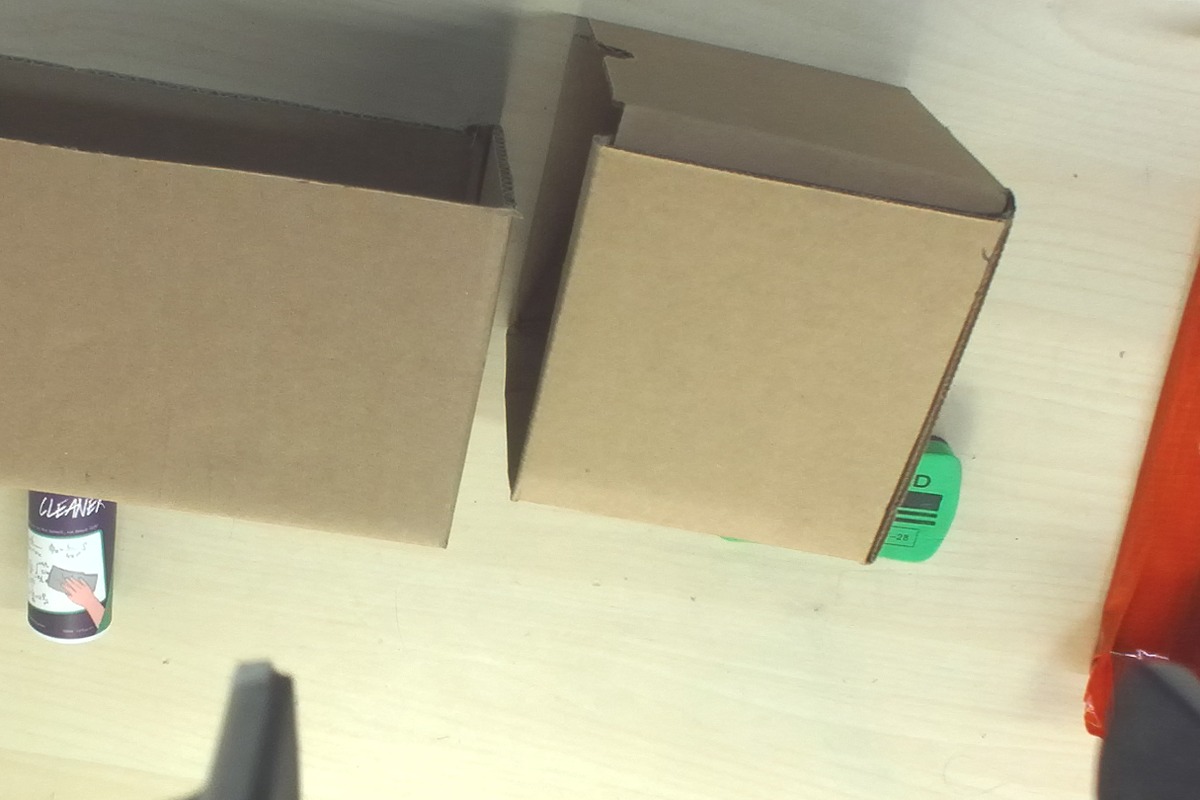}
    \caption{Supplies on Tray}
    \label{fig:scenes-supplies}
  \end{subfigure}
  \caption{The initial wrist-camera view of each task.}
  \label{fig:scenes}
\end{figure}

We perform evaluations on the Franka FR3 with a ZED-M wrist camera with known intrinsics and extrinsics. The perception pipeline uses Gemini-ER-2~\cite{team2025gemini} as the VLM, SAM2~\cite{ravi2025sam} as the segmentation model, and RecGen~\cite{zadaianchuk2026recgen} as the image-to-3D reconstruction model. The fully observed TAMP solver we use is cuTAMP~\cite{shen2024differentiable}, a GPU-parallelized TAMP solver.
Our illustrative manipulation settings are shown in Figure~\ref{fig:scenes}.

\begin{table}[t]
  \centering
  \small
  \setlength{\tabcolsep}{4pt}
  \caption{Success rate (successful trials / attempted trials) for each method on each scene.}
  \label{tab:results}
  \begin{tabular}{l ccc}
    \toprule
    & \multicolumn{3}{c}{SR $\uparrow$} \\
    \cmidrule(lr){2-4}
    Method & \makecell{Cubes on\\Bread} & \makecell{Teacup\\on Plate} & \makecell{Supplies\\on Tray} \\
    \midrule
    TiPToP~\cite{shen2026tiptop} & 2 / 10 & 3 / 10 & 0 / 10 \\
    \texttt{VUTAMP-UNIFORM}      & 8 / 10 & 3 / 10 & 2 / 10 \\
    \texttt{VUTAMP}              & 8 / 10 & 9 / 10 & 2 / 10 \\
    \ourmethod{}                 & 9 / 10 & 10 / 10 & 10 / 10\ \\
    \bottomrule
  \end{tabular}
\end{table}

In \textbf{Cubes on Bread}, the robot must pick up two stacked blocks and place both of them on the slice of bread. Moving the top block reveals the extent of the bottom block.

In \textbf{Teacup on Plate}, the robot has to move the blue cup from the table and to the flower-shaped plate. To do this, it must first take the occluding block off the cup.

The \textbf{Supplies on Tray} task features two cleaning supplies that are occluded by the boxes in the scene, which the robot has to look behind. The tray that the supplies have to be placed on also has collision obstacles that are not visible from the initial view.

\subsection{Results}
Our results are in Table~\ref{tab:results}. We report the success rate of each method on each task. We find that each layer of added task complexity motivates another component of \ourmethod{}. In the \textbf{Cubes on Bread} task, the bottom cube was occluded, which required a memory system to associate and update the initial bottom cube representation after revealing the full extent. TiPToP's failures on this task could be divided largely into grasp failures due to trying to place the robot's finger inside the body of the bottom cube, and placement failures due to trying to place the bottom cube onto a region of the toast that the top cube had already occupied. Because \texttt{VUTAMP-UNIFORM} was able to update its memories, it could outperform TiPToP.

In the \textbf{Cubes on Bread} task, there were only three objects in the scene and they were all task-relevant. The \textbf{Teacup on Plate} task introduces another layer of complexity by cluttering the scene with many task-irrelevant objects. An OBM that focused on greedily improving the worst mesh in the scene would miss the task-relevant objects, the teacup on the plate and the ones on the table the robot would pick up. By weighing the objects higher due to their proximity to the robot in its nominal plan, the objects are refined earlier by the OBM, leading to \texttt{VUTAMP} outperforming \texttt{VUTAMP-UNIFORM}.

In the \textbf{Supplies on Tray} task, the robot does not incidentally receive useful information while performing goal-oriented tasks. Unlike the preceding two tasks, where removing occluders was geometrically necessary to reach target objects, the cardboard boxes occlude the cleaning supplies visually but do not prevent the robot from reaching for those supplies. The two main failure modes for this task were grabbing an object about its longer axis (which seemed collision-free due to the cardboard box occlusion), or trying to place the objects on the red tray at an orientation that caused collision with the obstacles already there. In both cases, the robot had to plan observations which did not directly interact with the objects in the scene, but did serve to validate the robot's scene models. \ourmethod{} is the only method that does this and thus has the strongest performance on this task.

\section{CONCLUSION}
We presented \ourmethod{}, an approach to \textit{partially observable object rearrangement} that combines a persistent object-based memory with a deterministic TAMP solver augmented with information-gathering actions. Our tabletop experiments showed that both components were necessary in scenes with distractor objects, occlusion, and out-of-frame obstacles.

Our approach assumes that the initial observation supports a nominal plan that can guide information gathering. Integrating object search~\cite{zheng2023system, honerkamp2024language} could extend it to settings in which some objects are missing from the scene model. We currently use generative shape models conditioned on a single image. Models that incorporate multiple observations~\cite{li2026mv, zhang2026scene} offer a promising alternative, but their inference time scales poorly with the number of views. Selecting informative or diverse subsets of observations could help make these models practical within the OBM. Finally, reconstruction methods that combine RGBD observations with contact~\cite{bianchini2025vysics, saund2022clasp, smith2021active} could extend plan justification to include touch-based verification.

\section*{ACKNOWLEDGMENTS}
We gratefully acknowledge support from AFOSR grant FA9550-22-1-0249; from ARO grant W911NF-26-1-A329; from ONR MURI grants N00014-22-1-2740 and N00014-24-1-2603; from the MIT Siegel Family Quest for Intelligence; and from the Robotics and AI Institute.

AI Disclosure: Under the authors' direction, OpenAI Codex helped copyedit Sections~I--V and the TikZ code for Figure~\ref{fig:method-overview}. Anthropic Claude assisted with logging code for experiment assets. The authors take full responsibility for the paper.

\bibliographystyle{IEEEtran}
\bibliography{refs}
\end{document}